\documentclass[journal]{IEEEtran}

\usepackage[disable]{todonotes}

\usepackage{tabularx}

\usepackage{cite}
\ifCLASSINFOpdf
\else
\fi

\usepackage{subfig}

\ifthenelse{\boolean{true}}{
    \usepackage{eso-pic}%
    \AddToShipoutPictureBG*{%
        \AtPageUpperLeft{%
            \raisebox{-4\baselineskip}{
\makebox[\paperwidth]{\colorbox{yellow}{\begin{minipage}{0.85\paperwidth}\centering\small
                             {This work has been submitted to the IEEE for possible publication. Copyright may be transferred without notice, after which this version may no longer be accessible.}
            \end{minipage}}}}%
        }
        \AtPageLowerLeft{%
            \raisebox{3\baselineskip}{
\makebox[\paperwidth]{\fbox{\begin{minipage}{0.85\paperwidth}\scriptsize
                             {}
            \end{minipage}}}}%
        }
    }
}{}

\begin{document}
\title{I Do Not Understand What I Cannot Define:
Automatic Question Generation With Pedagogically-Driven Content Selection}

\author{{Tim Steuer,
        Anna Filighera,
        Tobias Meuser
        and Christoph Rensing}%
\thanks{The authors are with the Multimedia Communications Lab, Technical University of Darmstadt, Rundeturmstr. 10,
64283 Darmstadt e-mail:  \{tim.steuer,anna.filighera,tobias.meuser,christoph.rensing\}@kom.tu-darmstadt.de.}%
\thanks{Manuscript received XXXXX}}

\markboth{IEEE Transactions on Learning Technologies,~Vol.~XX, No.~X, XXXXXX~20XX}%
{Steuer \MakeLowercase{\textit{et al.}}: Automatic Question Generation With Pedagogically-Driven Content Selection}
\maketitle

\begin{abstract}

Most learners fail to develop deep text comprehension when reading textbooks passively.
Posing questions about what learners have read is a well-established way of fostering their text comprehension. However, many textbooks lack self-assessment questions because authoring them is time-consuming and expensive. Automatic question generators may alleviate this scarcity by generating sound pedagogical questions.
However, generating questions automatically poses linguistic and pedagogical challenges. What should we ask? And, how do we phrase the question automatically? We address those challenges with an automatic question generator grounded in learning theory. The paper introduces a novel pedagogically meaningful content selection mechanism to find question-worthy sentences and answers in arbitrary textbook contents. We conducted an empirical evaluation study with educational experts, annotating 150 generated questions in six different domains. Results indicate a high linguistic quality of the generated questions. Furthermore, the evaluation results imply that the majority of the generated questions inquire central information related to the given text and may foster text comprehension in specific learning scenarios.

\end{abstract}

\begin{IEEEkeywords}
Automatic Question Generation, Self-Assessment Technologies, Authoring Tools, Educational Technology
\end{IEEEkeywords}

\IEEEpeerreviewmaketitle

\renewcommand{\arraystretch}{1.3}

\section{Introduction}

\IEEEPARstart{R}{eading} is a fascinating process. It allows us to understand others' thoughts, regardless of time or space.
Whether the authors live in another city, reside in a different social community or have lived in a bygone era, we can preserve their thoughts and human knowledge through time and space through their texts.
Hence, learning by reading is ubiquitous and crucial for our education.
No matter if one wants to learn philosophy from René Descartes or quantum mechanics from Richard Feynman, there is usually a text to assist.
The need for excellent textbooks arises, especially in higher education.
More complex topics come to the fore, with no experts in the learner's peer group directly approachable.
Moreover, more abstract topics are less demonstrable.
While primary school pupils may still seek help from their parents or explanatory videos on the internet, these possibilities are hardly available to a master's student who wants to develop a deep understanding of a field.
In such cases, learning from textbooks may even be the only viable option.
Consequently, supporting learners' cognitive processes during the reading of textbooks can be crucial for their educational career.

Although the process of reading comprehension is not fully understood, research suggests that a variety of interconnected mental representations emerge during reading~\cite{mcnamara2009toward,best2005deep,graesser2010good}. According to a review by McNamara et al., many models assume a propositional network as the basic architecture for the mental representations ~\cite{mcnamara2009toward}. Readers store concepts and their relations in graph-like structures in their mind ~\cite{mcnamara2009toward}. According to McNamara et al. ~\cite{mcnamara2009toward} for expository texts, such as higher educational textbooks, a well-established model is the construction-integration model (C-I model) of text comprehension~\cite{kintsch1998comprehension} which distinguishes three interconnected mental representations: the surface model, the text-base and the situational model.
The surface model stores linguistic information on the reading material. In the text-base, linguistic knowledge is transformed into the propositions explicitly stated in the text. In other words, it is the semantic representation of the textual content. Finally, the situational model enriches the text-base's propositions with inferences and interlinks them with the readers' prior knowledge ~\cite{kintsch1998comprehension, mcnamara2009toward}.
Hence, well-connected and densely populated text-bases and situational models are crucial for text comprehension because they represent the semantic information needed to understand the text~\cite{rouet2002mining}.

A large body of research points to the inefficiency of learning by passively reading texts ~\cite{hamilton1985framework,anderson1975asking,rouet2002mining,duke2009effective} and the theories of reading comprehension can explain why passive reading is inefficient.
Passive reading typically results in incomplete and only poorly constructed mental models, thereby hindering text comprehension~\cite{rouet2002mining}.
In higher educational textbooks, the effect is more pronounced\cite{best2005deep}. Textbooks are inherently difficult, as the prior knowledge needed to construct valid mental representations is often conveyed by the very same textbook. Hence, learners may not rely on their common sense or everyday prior knowledge when building mental representation for textbooks, as they do for less complex topics or stories ~\cite{rouet2002mining}.
Consequently, fostering the construction of complete mental representations is vital for improving learning from scientific textbooks.
One way of scaffolding the construction of complete mental representations is to ask learners meaningful questions about what they have read.
A meaningful question asks about central information needed to understand the book's topics. It scaffolds the construction of a coherent mental model. For instance, given a text about \emph{The Structure of Atoms}, the question \emph{What is the nucleus of an atom comprised of?} is valuable whereas the question \emph{What was the first name Rutherford gave a proton?} may be well-formed, but not very helpful to understand the actual content of the text.
On a theoretical level, questions affect learners by urging them to re-validating their mental models based on the questions ~\cite{rouet2002mining}.
Moreover, questions reactivate propositions stored in the learners' memory during reading. The subsequent retrieval of them from the learners' memory favours permanent storage of the propositions in the long-term memory a phenomenon called testing-effect ~\cite{roediger2006test}.
In conjunction with these theoretical considerations, there is ample evidence supporting that posing questions about what learners read, increases their learning outcomes ~\cite{rouet2002mining,anderson1975asking }.
Studies have found an increase in near transfer learning outcomes, far transfer learning outcomes and better long-term memory retrieval ~\cite{hamilton1985framework,roediger2006test}.
Concerning near transfer, multiple studies show the benefits of questions. Learners can answer questions about facts better if they already encountered similar questions about these facts during reading ~\cite{anderson1975asking,hamilton1985framework,rouet2002mining}.
Regarding far transfer, studies found that learners are able to make better inferences about the text if they are asked comprehension questions during reading ~\cite{hamilton1985framework}.
Finally, multiple studies demonstrated that for long-term memory retrieval, learners remember more about texts when they answered questions about the text than reading the text multiple times ~\cite{roediger2006test,dirkx2014testing}. The testing effect becomes more robust over time, and learners with questions outperform passive readers significantly when asked to recall a text's contents~\cite{roediger2006test}.

\begin{figure}[tb]
\centering
\includegraphics[width=0.25\textwidth]{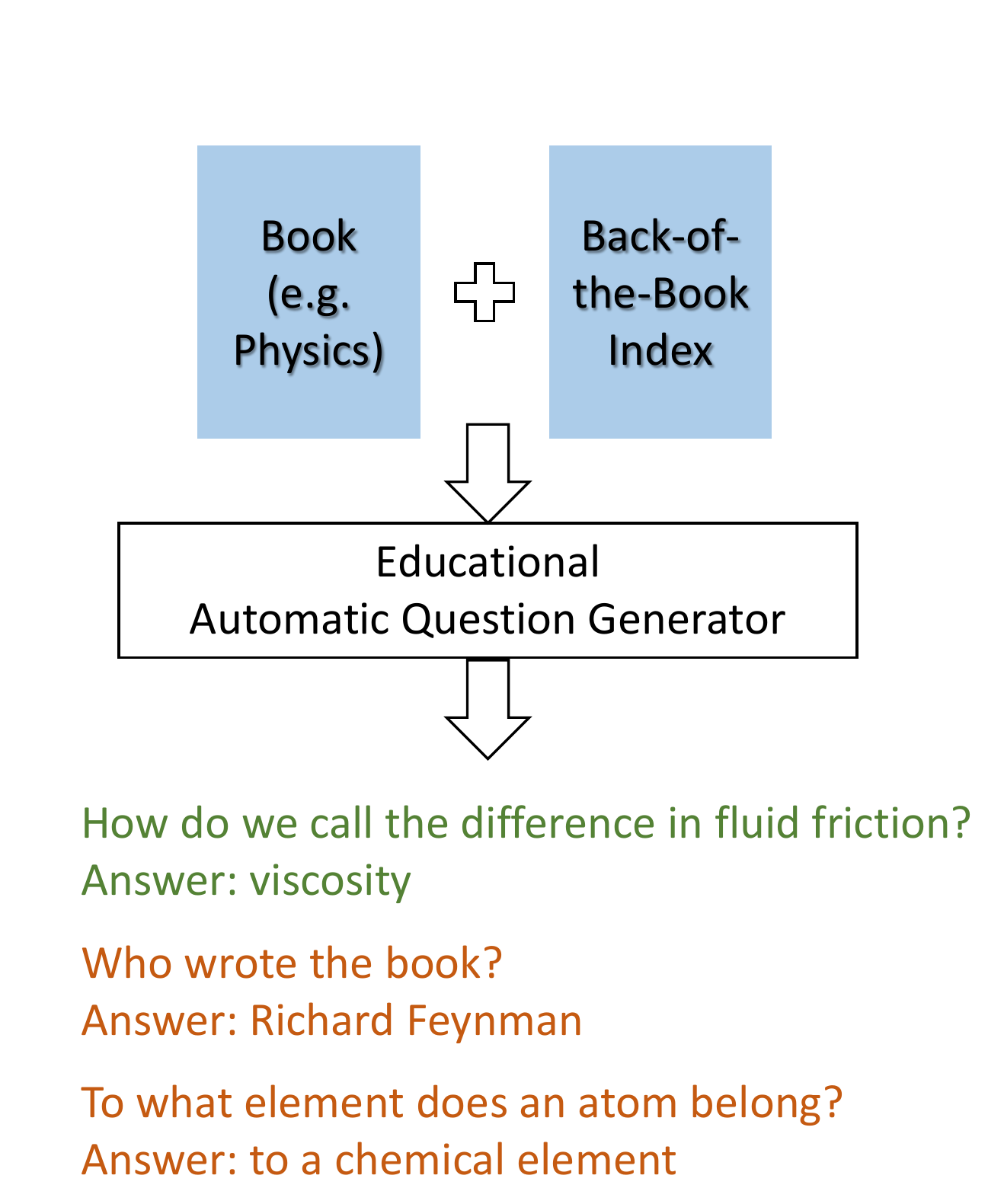}
\caption{Educational automatic question generation for textbooks, as discussed in this paper. Given a textbook and its back-of-the-book index, generate meaningful pedagogical questions about the text (green question). Current automatic question generation systems usually ask about arbitrary text content (orange questions).}
\label{fig_idea}
\end{figure}

Although questions positively affect the understanding of textbooks, most textbooks lack a sufficient number of questions about the text. Generating questions is additional work for authors, and most authors need time and training to come up with the right questions on their own~\cite{graesser2010good}. Furthermore, we can not rely on learners coming up with their own questions because they only pose questions infrequently and in a shallow manner~\cite{graesser1994question}.
As a result, textbooks are often hard to understand for learners, and they fail to develop a good understanding of the topics discussed. Consequently, we aim to foster readers comprehension of textbooks used in higher education via automatic question generation.

We investigate the following research question:

\begin{itemize}
\item[RQ]{To what extent can educationally meaningful questions about textbooks from higher education be generated automatically?}
\end{itemize}

We do not restrict the task to a specific domain and experiment with textbooks from various fields such as physics or sociology.
Our contributions to this challenging task include:

\begin{itemize}
\item{an educational automatic question generation (AQG) process for complete textbooks}
\item{a content selection approach grounded in educational theory}
\item{an expert annotation study investigating the AQGs linguistic and pedagogical qualities}
\end{itemize}

The proposed educational AQG process takes a textbook and its back-of-the-book index as inputs (Fig. \ref{fig_idea}).
It comprises a content selection and a textual question generation phase. During content selection, the process determines if a given sentence is meaningful for question construction and what sentence part may represent a valid answer. This information is used during textual generation to construct the actual question.
While recent works have shown that content selection for educationally valuable question generation is hard ~\cite{chen2019comparative} and that many non-educational systems fail to ask about pedagogically valuable contents~\cite{horbach2020linguistic}, our expert annotation study result suggests that the generated questions are linguistically-sound, central to the text and possess pedagogical value.

What follows is an introduction to the related work tackling content selection and textual question generation, focusing on the educational domain. In chapter three, the proposed educational AQG approach is presented. We define the technical problem and describe the process steps for content selection and textual question generation in detail. Chapter four reports the methodology and results of our evaluation study conducted with three educational experts. Chapter five discusses the evaluation study's findings and limitations concerning our research question. Finally, chapter six concludes the paper.

\section{Related Work}

Educational AQGs working with textbooks must at least have two different capabilities.
First, adequate content has to be selected for which meaningful pedagogical questions can be asked.
Second, the AQG must pose a grammatical sound and meaningful question concerning the selected content.
This distinction between \emph{content selection} and \emph{textual question generation} is intuitive and prevalent in the literature \cite{piwek2012varieties, du2017identifying, steuer2020remember}. Therefore, our review of the related work is split among these two problems.
However, note that content selection and textual question generation are not completely decoupled.
Depending on the generator's input requirements, the content selection may result in a small paragraph, a single sentence, or even a sentence and answer candidates in the sentence.
Consequently, not every content selector is combinable with every question generator.
We focus on the most influential works for our problem at hand (textbook question generation). For an exhaustive summary of educational AQG in general, we refer to the survey by Kurdi et al. \cite{kurdi2020systematic}. For a review of state-of-the-art neural question generation, we recommend the review by Pan et al. \cite{pan2019recent}.

\subsection{Content Selection}

The content selection comprises reducing the amount of text in a chapter to its most meaningful subparts useful for asking a question. We call this process context selection and depending on the connected generator's input format, a context corresponds to small paragraphs or single sentences.
Furthermore, it may also comprise the explicit marking of answer-candidates in the selected context sentences. We call this process answer selection.
Context selection was tackled through various text statistics\cite{huang2016automatic, kumar2015revup}, through methods stemming from extractive summarization\cite{chen2019comparative, rudian2020educational} or through text classifiers\cite{du2017identifying, wangqg}.

An example of statistic-based selection on reading comprehension texts was proposed by Huang et al. \cite{huang2016automatic}.
They showed that text statistics measuring the keyness, completeness and independence of sentences could already result in a good selection of sentences for language learners works well for short, easy and intermediate texts.
However, the authors limit their methodology to short reading comprehension texts for language learners.
Their approach does not transfer to more complex texts, as they often lack an optimal selection of keyness, and sentences are often interdependent.
Different text statistics, using topic distributions for ranking sentence by importance were proposed by Kumar et al.~\cite{kumar2015revup}, assuming that a few topically important coherent sentences will result in the generation of good questions. Yet, no evaluation of the approach was provided, and the authors' assumption will not always hold.
A non-central sentence may provide precisely the additional information for a topic, that is essential for the learner's understanding.

The transfer of extractive summarization methods to context selection has provided mixed results \cite{chen2019comparative, rudian2020educational}. In experiments with different data sets, no single best extractive summarization algorithm for content selection could be found \cite{chen2019comparative}. Although LexRank\cite{erkan2004lexrank} achieves the best scores in a comparison study of different algorithms on different datasets, the study results' are only a very rough estimate, as the study measures performance by generating questions via an upstream question generator. Consequently, the achieved BLEU-4 \cite{papineni2002bleu} scores depend directly on the study's question generator's quality and distorts the analysis. Another comparison study, based on German input texts derived from Wikipedia, uses educational experts for their evaluation\cite{ rudian2020educational}. In a comparison of nine algorithms, they found LexRank inferior to Edmundson's text summarization algorithm\cite{edmundson1969new}.
Both studies indicate that text summarization algorithms can provide a promising starting point for content selection, but must be fine-tuned to the target domain. For example, both Edmundson's algorithm and LexRank require the appropriate frequency matrices for the given domain to provide useful content selection.

Finally, classifier-based methods have been applied to context selection.
They either classify a sentence or a part of a sentence as question-worthy by applying typical pattern matching algorithms like logistic regression or neural-networks \cite{du2017identifying, wangqg}.
It is demonstrated that these methods yield convincing results and achieve good F1-measures, precision and recall scores on the respective test sets \cite{du2017identifying, wangqg}. Their main weakness is that the classifier learns whether a sentence constitutes valuable information implicitly from the training data set.
In other words, what constitutes question-worthy information is not determined a priori by pedagogical considerations but learnt from a corpus.
It is unclear why certain information is question-worthy according to the classifier, making it hard to determine why the algorithm selects sentences.
Likely, the learnt selection strategy is not driven by pedagogical-considerations but by surface-level text features.
Hence, this implicit knowledge is not necessarily transferable to texts with dissimilar structure or content.
Ergo, although their performance on their respective test sets is high, we assume a direct usage on arbitrary textbooks will yield worse results.

\begin{figure*}[thb]
\centering
\includegraphics[width=\textwidth]{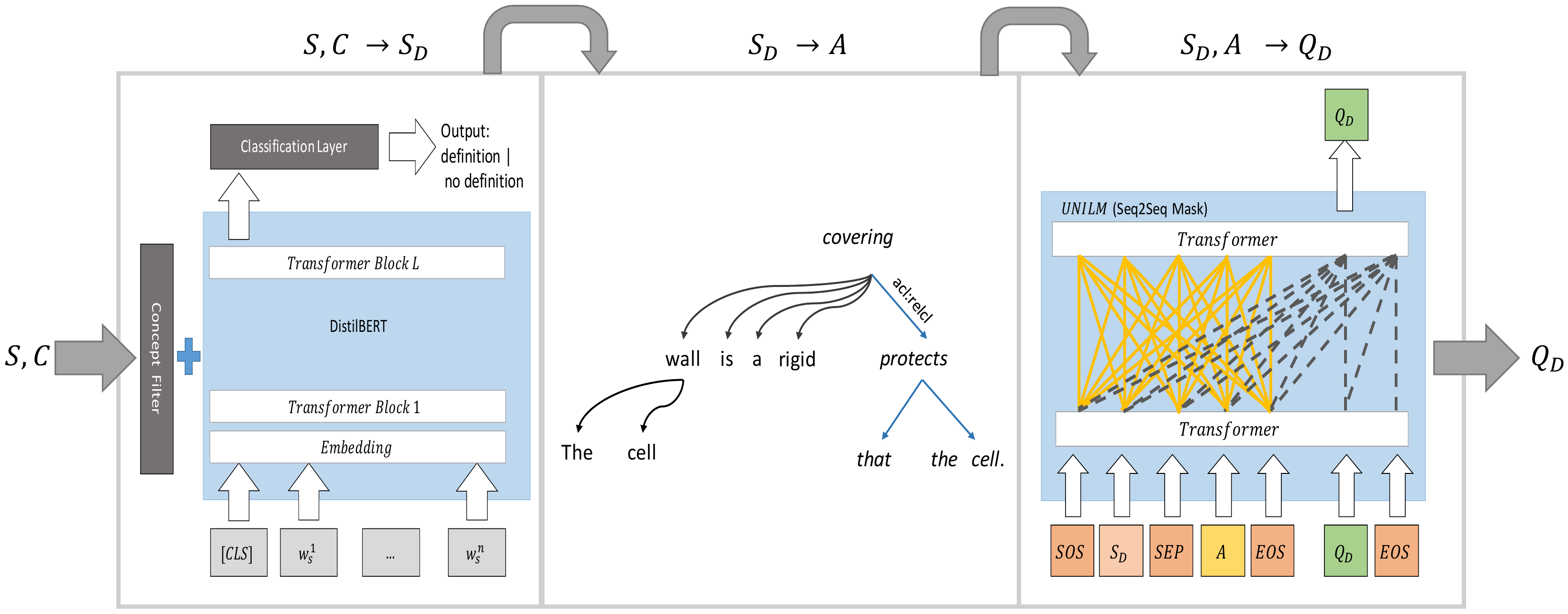}
\caption{The overall architecture of our system. Left, the context selection step using a keyword filter and a DistilBERT \cite{sanh2019distilbert} classifier. Every \(w_S^i \) is a word in the given input sentence \(S\). Center, the answer selection step based on semantic graph matching \cite{chambers2007learning} on the dependency graph of a given definition. In the example, the italic words are extracted because they constitute a relative clause (other relations are omitted for readability). Right, the UNILM transformer \cite{dong2019unified}, which given answer and definition, generates a factual question. It applies a bidirectional mask to the given definition and answer and a left-to-right mask to the question under generation. The inputs \(A\) and \(S_D\) are represented as sequences of words. The sequence \(Q_D\) is autoregressively generated.}
\label{fig_overall_architecture}
\end{figure*}

Besides, context selection, many generators also explicitly need the answer the question should ask for. Therefore, the second step of content selection comprises answer candidate selection inside a given span of text. Multiple approaches have been presented. Classification-based methods have been applied \cite{willis2019key,becker2012mind} with similar strengths and weaknesses as in context selection. Furthermore, various parsing and tagging technologies were used to select the answer inside sentences. It has been shown that semantic role labelling and dependency parsing can be used to find meaningful answers in educational settings \cite{mazidi2016infusing}.
Moreover, matching techniques over the dependency graph, such as Semgrex patterns \cite{chambers2007learning} may also be applied to improve the meaningfulness of generated questions compared to answer-unware generators \cite{steuer2020remember}. One advantage of these methods is that they do not require training data and can be applied to a wide range of texts. However, the disadvantage is that the semantic meaning of the answer is often disregarded.

\subsection{Textual Question Generation}

The literature distinguishes three main methods for generating questions from textual inputs.
First, rule-based approaches transform a declarative sentence into a question via grammatical transformations (e.g. \cite{huang2016automatic}).
While working well for simple sentences, such systems often fall short when working with arbitrary texts and complex sentences \cite{ du2017learning}.
Second, template-based approaches have fixed templates such as \emph{``What is the advantage of X?''} and fill the variables in those templates (e.g. \cite{mazidi2016infusing}).
Templates work well in limited domains if there are enough to cover the language phenomena in a given text.
Reaching high coverage on arbitrary textbooks is challenging because of the different writing styles, and vocabulary in different domains \cite{mazidi2016infusing}.
Furthermore, each template must be created manually, and the resulting questions tend to sound repetitive.
Third, neural question generators (NQGs) use statistical language models to transpose declarative inputs into questions (e.g. \cite{wang2018qg}).
In the educational sector, rule-based and template-based systems are the most widespread \cite{kurdi2020systematic}.
Outside the educational sector, neural question generators outperform their template-based, and rule-based competitors \cite{du2017learning} in the intrinsic evaluation of the grammaticality, the expressiveness and the naturalness of the generated questions.
While first NQGs where simple sequence-to-sequence models transposing a single input sentence\cite{du2017learning}, recently more complex neural architectures have been investigated.
Answer-aware architectures (e.g. \cite{kim2019improving}), and complex pretrained transformer-based architectures (e.g. \cite{dong2019unified}) have lead to further performance improvements.
First experiments with such neural architecture in the educational domain are promising, showing that such methods outperform rule-based methods on educational datasets measured by automatic metrics\cite{chen2018learningq}. Moreover, they may generate deeper comprehension questions by generating seemingly plausible and fluent text on the topic, on which learners must position themselves by concluding what they read \cite{steuer2020exploring}.

\section{Approach}

In the following, we describe our educational AQG system. First, we further specify the problem of educational useful question generation and outline our approach's core ideas.
Next, we discuss the different parts of the educational AQG in detail.

\subsection{Problem Definition}

We argue that explicit pedagogical priors must govern the content selection to pose educationally meaningful questions.
Hence, we make explicit pedagogical assumptions about what constitutes meaningful content for question generation. We assume that definitions, in particular, are vital for building a coherent text-base. After all, a coherent text-base contains propositions about the text's most important concepts~\cite{rouet2002mining}, which are introduced precisely through definitions. Furthermore, this explicit pedagogical assumption enables us to select the expected answers for the questions generated based on definitions' specific language properties. Thus, although many possible questions may be constructed for a given definitional sentence, of which some are not useful for learning, we can often determine the answer to the question to be generated beforehand. Hence, alleviating the chance of meaningless questions being generated. Based on these considerations, we derive the following educational AQG task.
Given the sentences in a textbook \(S\) and the crucial concepts of the book \(C\), generate meaningful educational questions \(Q_D\).

\begin{equation}
ask: S \times C \rightarrow Q_D
\end{equation}

Every question \(q \in Q_D\) targets factual knowledge transmitted by a definitional sentence of the given text.
Hence, not all sentences of the book are used to generate questions.
Furthermore, each question is assignable to the source paragraph of its source sentence and is answerable from the information provided there.
Consequently, readers can answer the questions by revalidating their text-base.
The set \(S\) is automatically derived by sentence tokenizing the textbook content. The set \(C\) is currently manually extracted from the back-of-the-book index but may also stem from automatic keyphrase extraction if no index is present or from user preferences to allow personalized question generation.
The function \(ask\) is composed of a context selection, an answer selection and a textual question generation step (Fig. \ref{fig_overall_architecture}).

\subsection{Context Selection}

The set \(S\) comprises every sentence of the textbook.
Thus, we determine a subset \(S_D\) of these sentences for which, based on our pedagogical considerations, it is reasonable that they will be question-worthy.

\begin{equation}
contextSelect: S \times C \rightarrow S_D
\end{equation}

We consider a sentence \(s \in S_D\) question-worthy if it encompasses at least one keyphrase \(c \in C\) and also has definitional character. Note that limiting the context selection to single sentences instead of small paragraphs is restrictive.
Yet, we treat sentence boundaries similarly to Spala et al. \cite{spala2019deft}, who have developed an annotation scheme for definitions and consider cross-sentence information as only supplemental.
The context selection consists of a keyword filtering over \(S\) and a definition classifier for all remaining sentences.

For keyword filtering a case-insensitive regular expression-based filter is applied.
It preserves sentences containing single-word or multiword concepts listed in the back-of-the-book index.
Whereas filtering according to keywords is straightforward, definition classification is more complex as sentences can define concepts differently.

\begin{figure*}[tb]
\centering
\subfloat[ROC curve for the BERT-based classifier on the DEFT dev set.]{\includegraphics[width=0.5\textwidth]{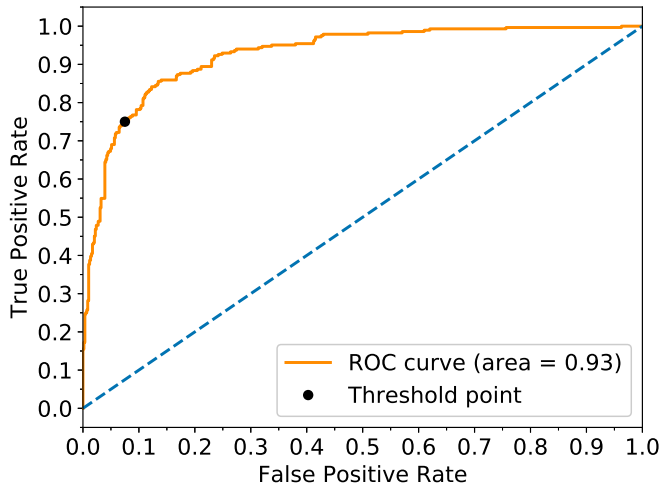}%
\label{fig:roc}}
\hfil
\subfloat[The question count decreases slowly with increasing classification threshold on the observed books in various domains: anatomy, biology, chemistry, physics, psychology and sociology.]{\includegraphics[width=0.5\textwidth]{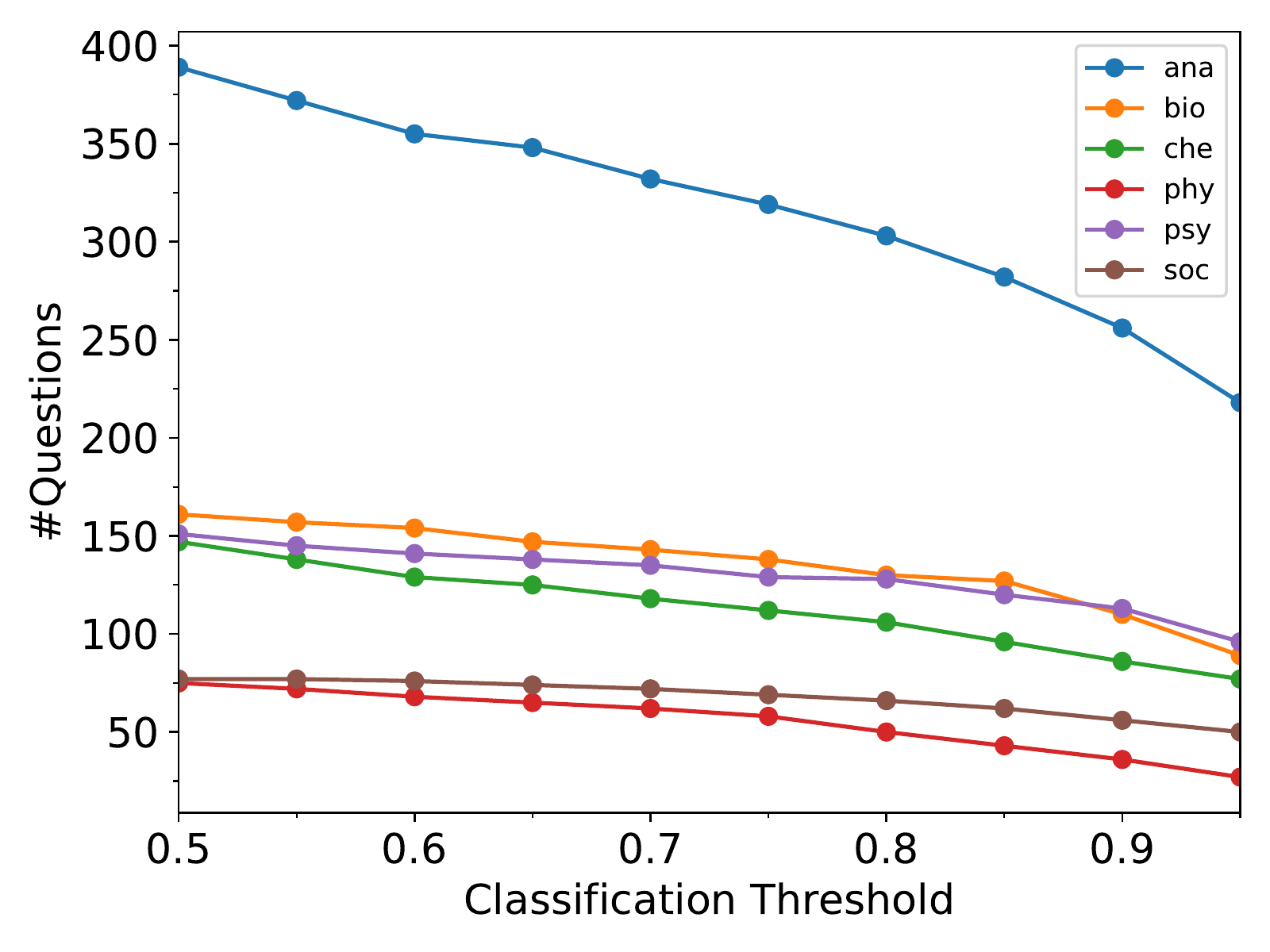}%
\label{fig:question_count}}
\caption{Analysis of the effect of different classification thresholds on the true-positive and false-positive rate and the amount of questions generated. }
\label{fig_context_select}
\end{figure*}

On the one hand, some definitions comprise explicit linguistic cues such as the verb phrases \emph{``refers to'}' or \emph{``is defined as ``} \cite{spala2019deft}.
On the other hand, many definitions do not follow this strict form. They lack explicit linguistic cues such as the sentence: \emph{``This class of upwardly mobile citizens promoted temperance, or abstinence of alcohol.''} which defines \emph{temperance} as the \emph{absitenence of alcohol} \cite{spala2019deft}.
Therefore, although it seems possible to form simple rules for definition classification, rules are ineffective, overlooking many valid sentences \cite{spala2019deft}.

As a consequence of rule-based approaches being inferior, recently, a large corpus (DEFT) \cite{spala2019deft} has been proposed, comprising around 30,000 sentences from legal documents and textbooks of multiple domains, of which around 11,000 are definitions.
The SemEval-2020 Task 6 \cite{spala2020semeval} investigates how to use the proposed corpus for machine learning-based definition extraction approaches which often generalize better than handcrafted rules.
We make use of this corpus and built a machine learning-based definition classifier.
While the DEFT corpus allows more complex modelling, such as tagging the concept to be defined (the definiendum), we use it to train a simple binary classifier to discard non-definitional sentences.

On first sight, our context selection thus resembles the classifier-based approaches from related work.
However, compared to previous work, we rely on explicitly grounding the content selection with pedagogical priors.
The proposed classifier does not learn implicitly what constitutes a question-worthy sentence.
Instead, we start with explicit pedagogical assumptions of what is question-worthy and what is not and encode these explicit assumptions into a classifier via the training data.
Consequently, we expect a more meaningful context selection and improved generalization capabilities. The proposed classifier will generalize to unseen data as long as the way definitions are expressed in those texts are not entirely different. Because the DEFT corpus contains already rather many formulations expressing definitions, this should usually be the case. In contrast, implicitly learnt content selectors are more likely to fail on unseen data because it is unclear if the selection criteria they learnt hold for different data distributions.

A DistilBERT model \cite{sanh2019distilbert} trained on the DEFT corpus's training split was chosen for the classifier's concrete implementation.
DistilBERT has shown almost state-of-the-art performance on many natural language benchmarks while also having small memory and computational costs \cite{sanh2019distilbert}.
Furthermore, variants of BERT-based transformers have shown promising results on the DEFT definition classification task \cite{spala2020semeval}.
Consequently, we opt for DistilBERT to reduce our pipeline's overall resource footprint.
We start with the pre-trained cased model and finetune it for three iterations without hyperparameter optimization and a maximum sentence length of 256 words on the DEFT corpus training set.
Our final model achieves the following statistics on the DEFT corpus classification test set: \(precision=0.78\), \(recall=0.78\) and \(f_1=0.78\).

With this model, most definitions can already be classified correctly, making it a good building block for the context selection.
However, false-positives and false negatives are inherent in machine learning-based classification, especially if we apply the model to previously unseen out-of-distribution data.
In other words, if we want to apply the model to arbitrary textbook content from various domains, we will see cases where it either classifies an actual definition as non-definitional or a normal sentence as a definition.
We anticipate that such errors are unavoidable and adapt our question generation accordingly.

Human-computer interaction studies relate user acceptance of such errors to the cost associated with the error, which is task-specific \cite{kocielnik2019will}.
For example, it is more acceptable for users that a scheduling assistant wrongly detects appointments in emails than that it overlooks appointments that take place.
The reason is, ignoring the false appointment is easy for users, whereas missing an appointment comes with high costs \cite{kocielnik2019will}.
In such cases, the system should favour recall over precision.
A similar design consideration affects the definition classification in the context selection step.
When reading a complex text, readers' cognitive load is already high.
Furthermore, readers tend to assign higher importance to concepts they are asked about \cite{duke2009effective}.
Hence, thinking about unimportant questions is associated with high costs.
Consequently, we emphasize precision over recall, reducing unwanted questions by also reducing the total amount of questions generated.

A suitable classification threshold yielding high precision and acceptable recall can be determined by analyzing the true-positive and false-positive rate given different classification thresholds as visualized by the ROC plot (Fig. \ref{fig:roc}).
We plot the true-positive and false-positive rate under varying decision boundaries on the dev set of DEFT, to determine a good trade-off between true-positives and false-positives.
Having a classification threshold of \(0.7\) results in a high true-positive rate of \(0.75\) and an acceptable false-positive rate of \(0.08\).
In other words, we correctly identify \(75\%\) of all actual definitions in the test set as such, while only incorrectly identifying \(8\%\) of normal sentences as definitions.
Shortly after, the ROC curve begins to flatten, so that in relative terms, fewer true-positives are found by accepting more false-positives.
Thus, from a precision-recall point of view, we shall set the classification threshold of our definition classifier to \(0.7\).

Yet, the classification threshold naturally affects the number of questions generated per book.
If a precision-oriented classification threshold leads to too few questions generated, it is useless.
Hence, we need to verify the impact of the precision-oriented threshold on the number of questions generated before fixing it.
We examined how many questions will be generated for the first three chapters of textbooks from different domains under different given decision boundaries (Fig. \ref{fig:question_count}).
We start with a threshold of \(0.5\) and increase it in steps of \(0.05\) indicating a threshold geared towards higher precision.
The number of questions generated decreases only slowly with growing classification threshold.
In contrast, the ROC curve is increasing steeply, showing that high decision boundaries already yield good classification performances for the positive case.
In other words, the classification accuracy for the positive class improves disproportionately to the decrease of generated questions.
Consequently, it can be said that most classifications in textbooks are relatively unambiguous, and increasing the classification threshold does not have a drastic negative effect on the number of questions generated.

\subsection{Answer Selection}

The result of context selection is the set \(S_D\) of sentences defining key concepts of the text.
Next, the answer selection step maps every \(s \in S_D\) to one answer phrase \(a \in A\) or to an empty answer, where \(A\) is the set of all selected answers. Every \(a \in A\) comprises a subset of the words in \(s\) and must not be empty. If the answer selection results in an empty answer, the corresponding \( s \in S_D\) will not be considered any further.

\begin{equation}
answerSelect: S_D \rightarrow A
\end{equation}

Whereas AQGs exist that combine the answer selection and the textual generation of questions into one neural network, decoupling these steps has advantages.
It allows setting fixed pedagogical priors independent of any training data.
For most sentences, multiple questions may be formed, especially if we look at purely grammatical, not necessarily useful questions.
In other words, most sentences allow asking for pedagogically meaningless information.
If we leave answer selection to a learning algorithm, what constitutes a pedagogically meaningful answer will be implicitly defined by the data.
Hence, to have a chance to influence the questions generated for a sentence, identifying the expected answer beforehand is of great importance.

Every \(a \in A\) is a phrase with two properties to constitute a valuable and pedagogical meaningful answer.
First, the selected phrase must convey meaningful complete semantic information allowing the formation of a question. To illustrate, if we have sentence \emph{``The cat is a small furry animal liking humans.''} the word sequences \emph{``is a'' } or \emph{``furry animal liking''} may not be selected as an answer candidate, because one can hardly questions about these without asking for purely linguistic information (e.g. \emph{What is the second and third word in the sentence?}).
Second, as the inputs are definitions, the answer candidates should describe the definiendum's characteristics or should ask for the definiendum itself.
If asked about the defining characteristics, the reader must recall and associate them with the definiendum.
The recall is cognitively demanding and reactivates the definitions propositions in the mental model.
If given all characteristics and asked for the definiendum, fewer cognitive resources are required. Nevertheless, the definiendum's explicit wording is reactivated in the mental model, which is crucial for efficient communication about what has been learnt.
\todo{sind alle tables am richtigen spot}

The context selection likely implies that one of the noun phrases in the input sentence constitutes a definiendum.
Accordingly, selecting sentence parts that modify the noun phrases presumably yields the characteristics that illustrate the definition.
A typical way of doing such a modification in a language is using adjectival clauses such as relative clauses.
In everyday language they are often used to add important information about the noun phrase, e.g. in the sentence: \emph{``The carpet that you bought last year is beautiful''} the adjectival clause adds information.
This is also true for many definitional sentences in which they either introduce the definiendum, e.g. in the sentence\emph{``Intermediates of dsRNA, called replicative intermediates are made in the process of copying the genomic RNA.''} or specify important characteristics of the definiendum \emph{``A variable is any part of the experiment that can vary or change during the experiment''}.
Furthermore, adverbial clauses sometimes yield valuable information about definitions, e.g. when combined with a gerund as in the sentence \emph{``The cell wall is a rigid covering that protects the cell, provides structural support, and gives shape to the cell.''}\footnote{Example texts adapted from OpenStax, Biology 2e, OpenStax CNX. Mar 28,2018.}
The clausal structure of both grammatical constructs also indicates the representation of a complete semantic piece of information.
Therefore, we aim to extract adjectival and adverbial clauses from sentences by applying dependency parsing and semantic graph matching (see Fig. \ref{fig_overall_architecture} middle).

\begin{table}[t]
\caption{Coverage of the Semgrex patterns over the dev split of the DEFT corpus only considering actual definitions. We achieve an average coverage of around \(79\%\). }
\label{table_pattern_count}
\centering
\begin{tabularx}{0.5\textwidth}{>{\hsize=0.87\hsize}X >{\hsize=0.7\hsize}X >{\hsize=0.65\hsize}X >{\hsize=0.58\hsize}X >{\hsize=0.745\hsize}X}
\hline
Total Definitions & Without Matching Pattern & Adjectival Clause & Adverbial Clause & Direct Object  \\
\hline
 284 &   60 (21\%)& 146 (51\%)& 33 (12\%)& 45 (16\%) \\

\hline
\end{tabularx}
\end{table}
In dependency parsing, sentences are parsed into an acyclic directed graph whose nodes represent words and whose edges represent grammatical relationships.
The graph of a sentence represents the hierarchical relationship between the sentence's individual words.
The relationships expressed by the edges cover, for example, subject relationships, object relationships, or adjectival and adverbial clause relationships.
Since natural language is complex and very expressive, dependency parsing is usually addressed with statistical models.
They are often accurate but sometimes result in dependency graphs with faulty relations.
Yet, dependency parsing is a valuable tool to get additional linguistic information about a sentence.
Furthermore, it may be combined with semantic graph matching (Semgrex) \cite{chambers2007learning}.
A Semgrex pattern is similar to a regular expression.
It allows specifying a subgraph in the given dependency graph by matching the labels of links or nodes inside the graph.
For example, the subgraph of the sentence's subject, or the subgraph that serves as a relative clause.
Both tools combined allow us to extract answer candidates effectively.

We apply Stanfords CoreNLP Dependency Parser \cite{chen2014fast} to transform each \(s \in S_D \) into the corresponding dependency graph.
Next, we use four different semgrex patterns to extract subgraphs related to adjectival clauses and adverbial clauses.
Moreover, if we find more than one subgraph to extract in a single sentence, we heuristically opt for the subgraph with the most words, assuming it contains the most information.
If we do not find a subgraph to extract, we fall back to extract the sentence's direct object, assuming that even non-perfect answer candidates may still yield valuable questions.

Ideally, the selected patterns would apply to all definitions allowing the extraction of an answer for every \(s \in S_D\).
But, language is complex, and it is possible to form definitions without using the discussed grammatical constructs.
Yet, our approach is easily extendable and other, more complex, patterns could be added.
However, before we evaluate our system, we have to ensure sufficient coverage of the discussed patterns on a wide variety of texts.
Hence we analyzed the coverage of our patterns on the DEFT corpus dev split which comprises \(284\) definitions.
Every match of a pattern on a definition was counted, if multiple patterns match a single definition, only the first match was counted.
The proposed patterns cover around \(79\%\) of the definitions (see Table \ref{table_pattern_count}).
Over \(63\%\) comprise adjectival or adverbial clauses, whereas our fallback direct object heuristic covers \(16\%\).
We deemed this sufficient coverage to continue with an empirical evaluation of the system.

\subsection{Textual Question Generation}

In the final step, the goal is to generate the actual question text.
We generate one question for every pair of a definitional sentence and its answer candidate.
Hence our final system generates \(|A|\) questions.

\begin{equation}
generate: S_D, A \rightarrow Q_D
\end{equation}

Every \(q \in Q_D\) is a question asking about a definition. The following properties are useful for the generated questions.
First, the question generated must have high linguistic quality.
That is, must be free of language errors and read naturally.
Second, it must be answerable given the previously selected answer candidate.
To fulfil these requirements, we rely on the UNILM transformer\cite{dong2019unified}, which is a state-of-the-art neural question generator.

The UNILM model is a subword-based language model initialized with the BERT \cite{devlin2019bert} weights.
It applies three different attention masks during training and learns on multiple tasks at once.
The model is pre-trained on a large corpus comprising the English Wikipedia and the BookCorpus.
The model learns unidirectional, bidirectional, and sequence-to-sequence language modelling during pre-training through the different attention masks.
Hence, it can be finetuned for the corresponding natural language understanding or natural language generation task while at the same time, leveraging knowledge gained from the other attention masks.
For the AQG task, the sequenc- to-sequence mask is most important.
Here, the model receives one or multiple input sequences on which it attends bidirectionally (e.g. the definitional sentence and the answer candidate).
Next, it generates an output sequence, based on the given input sequence(s) autoregressively.
Because no right context exists, when generating a sentence word-by-word, the output sequence generation attends only left-to-right.
Consequently, this sequence-to-sequence attention mask, combines the strengths of bidirectional mask in typical NLU models (e.g. DistilBERT \cite{sanh2019distilbert}) and of autoregressive language models (e.g. GPT-2 \cite{radford2019language}).

One such sequence-to-sequence task is automatic question generation, for which UNILM has successfully been finetuned \cite{dong2019unified}.
It achieved close to state-of-the-art results in terms of BLEU-4 \cite{papineni2002bleu} scores (\(23.75\)) on the SQuAD benchmark dataset \cite{rajpurkar2018know}.
Furthermore, UNILM has been used in an educational setting before, where it generated questions with high linguistic quality \cite{steuer2020remember}.
Finally, the authors already provide a readymade snapshot for AQG, allowing more comfortable usage of the model for textual question generation.

We concatenate every definitional sentence and the corresponding answer candidate separated by the \emph{[SEP]} token into one file.
The final questions are generated using UNILM with the provided AQG snapshot and a batch size of 64.
At the end of our educational AQG pipeline, our result set comprises the triplets \( [(s,a,q) |\ s \in S_D, a \in A, q \in Q_D\)].
We map every triplet back to the initial paragraph of the textbook.
The questions may now be rendered together with the book.
Furthermore, the context sentence and answer candidate may be shown to readers as reference answers for the generated question.

\section{Evaluation Study}

In the following, we first introduce our research question and the reasoning for conducting an empirical evaluation study\footnote{the annotated data is available at https://github.com/t-steu/deft\_aqg/tree/master}. Next, we discuss the study's methodology, procedure, and reliability. Next, we discuss the general dataset statistics. Last, we provide a detailed look at the study's results.

\subsection{Research Question}

We aim to investigate the following research question in an explorative study:

\begin{itemize}
\item[RQ]{To what extent can we generate educationally meaningful questions about textbooks from higher education automatically?}
\end{itemize}

An educationally meaningful question possesses two properties.
First, it is linguistically sound.
Readers can understand the question directly, it reads naturally and is free of language-errors.
Second, it is pedagogical meaningful.
Hence, the question is related to the given material and can be answered.
To construct an answer, readers must use their common sense knowledge and use the information provided in the reading material.
Additionally, answering the question forces readers to think or recall important concepts and their characteristics, fostering their mental representations.

Neither the linguistic quality nor the pedagogical quality of a question can be measured by automatic means.
Although metrics such as BLEU or ROGUE are often used to estimate the linguistic quality of generated texts, they only infrequently correlate with actual human judgements \cite{liu2016not}.
Hence, the investigation of the given research question requires an empirical evaluation study.
We conduct the study with expert annotators having a background in educational sciences. They can best judge the pedagogical quality of the generated questions.
We aim to measure the quality characteristics of the generated questions concerning the discussed properties of educationally meaningful questions.

\subsection{Methodology}

\begin{figure}[tb]
\centering
\includegraphics[width=0.4\textwidth]{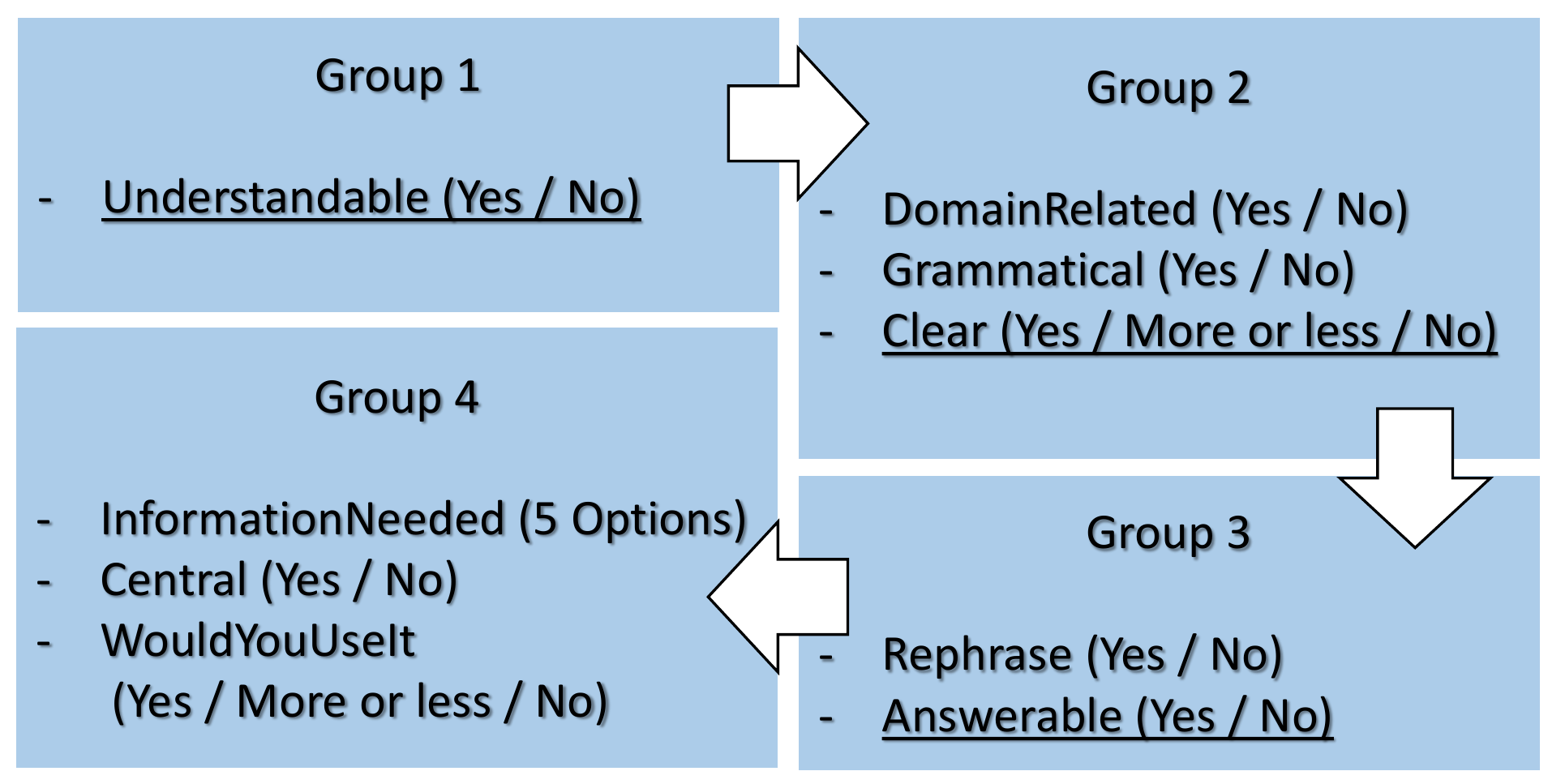}
\caption{The hierarchical annotation scheme applied. Every underlined question stops the annotation process if answered no. }
\label{fig_scheme}
\end{figure}

We conduct an empirical evaluation study with three educational experts.
Each of our experts has completed at least a two-year educational training. Their average age is 29, and all of them speak English on the C1 level according to the \emph{Common European Framework of Reference for Languages}.
Every expert judges the quality of the generated questions given the annotation scheme developed by Horbach et al. \cite{horbach2020linguistic}. The scheme explicitly aims to measure the generated questions' linguistic and pedagogical appropriateness (see Fig. \ref{fig_scheme}).
It comprises four groups of evaluation items measuring different quality characteristics of the generated questions:
general understanding, linguistic appropriateness, answerability and pedagogical appropriateness.
It consists of nine items, of which six are binary (yes/no), two have three answer choices, and one has five answer choices.
The scheme considers that for certain answer constellations it does not make sense to answer all evaluation items.
For this reason, there is an item in each of the groups, which, if answered \emph{"No"}, automatically marks the other items as \emph{``not applicable''} (underlined in Fig. \ref{fig_scheme}).
For example, if an annotator indicates in group three that the question cannot be answered, no more items from group four will be asked.
The hierarchy has two advantages: the evaluation categories' scale distributions are not distorted by ratings of generated questions that are not rateable, and the working load of the annotators is slightly reduced.

\begin{table}[tb]
\caption{The inter-annotator agreement of the three experts on the 150 questions of our evaluation study. The Krippendorff's \(\alpha\)'s are calculated on all 450 observations. The confidence interval (CI) of Krippendorff's \(\alpha\) is estimated via the boostrap.}
\label{table_iaa}
\centering
\begin{tabularx}{0.5\textwidth}{>{\hsize=1.5\hsize}X >{\hsize=0.5\hsize}X >{\hsize=0.6\hsize}X >{\hsize=0.5\hsize}X >{\hsize=0.5\hsize}X >{\hsize=0.8\hsize}X}
\hline
Item & \%-Agree\-ment & Krippen\-dorff's \(\alpha\) & CI Lower Bound & CI Upper Bound & Most Frequent Class\\
\hline
understandable & 0.81 & 0.35 & 0.29 & 0.40 & 0.83\\
domainRelated& 0.74 & 0.28 & 0.23 & 0.32 & 0.78\\
grammatical & 0.70 & 0.30 & 0.26 & 0.33 & 0.73\\
clear & 0.60 & 0.25 & 0.22 & 0.28 & 0.64\\
rephrase & 0.53 & 0.19 & 0.16 & 0.23 & 0.57\\
answerable & 0.67 & 0.22 & 0.18 & 0.26 & 0.73\\
informationNeeded & 0.42 & 0.18 & 0.15 & 0.21 & 0.39\\
central & 0.57 & 0.26 & 0.22 & 0.29 & 0.58\\
wouldYouUseIt & 0.41 & 0.16 & 0.13 & 0.19 & 0.41\\
\hline
\end{tabularx}
\end{table}

To generate our questions for the study, we use three chapters of six textbooks from different domains: anatomy, biology, chemistry, physics, psychology and sociology published by the OpenStax non-profit corporation.\footnote{openly available at openstax.org}
We opt for these books because they are a comprehensible introduction to the fundamental undergraduate topics in the respective fields and are written and peer-reviewed by experts.
Although the books are available as PDF, we do not automatically extract the content.
Automatic PDF text extraction would introduce erroneous texts, e.g. due to hyphenation, embedded fonts or other PDF specifics.
Thus, we manually extract the chapters' contents and back-of-the-book indexes based on the books' online version.
Furthermore, we ignore any exercise and any special purpose sections (e.g. figure captions or table headings) and only consider the main content.
We generate questions given each textbook's extracted data using our educational AQG approach.
Next, we apply stratified sampling to sample 25 of the generated questions from each of the six books. Hence, we select almost the same number of generated questions from every book randomly. The educational experts will annotate the resulting 150 questions.

Every expert annotates all of the 150 questions independently, in random order to alleviate order effects.
Annotators receive six questionnaires, each comprising around 25 questions.
Before every questionnaire, textual instruction on applying the annotation scheme, including good and bad examples was given similar to the instruction in the original study by Horbach et al. \cite{horbach2020linguistic}. The instruction could also be consulted through a help menu while annotating the data.
During the annotation of the data, the experts first read the textbook's corresponding paragraph before seeing one or multiple generated questions concerning the paragraph.
They judged the question on all of the annotation schemes dimensions simultaneously.

The inter-annotator agreement (IAA) between the three annotators is reported in Table \ref{table_iaa}.
We excluded one additional annotator from the study, due to systematic violation of the annotation scheme interpreting the clarity and answerability items incorrectly.
We report percentage agreement and Krippendorff's \(\alpha\) as measures of IAA for all data points, handling\emph{``not applicable''} as an extra answer choice.\footnote{IAA was computed using NLTK 3.4.5}
We also estimate confidence intervals for Krippendorff's \(\alpha\)'s using bootstrap sampling.
Similar to Zapf et al. \cite{zapf2016measuring}, we draw \(b=1,...B\) random samples (\(B=1000\)) of size \(N=1000\) with replacement from our data.
Every observation in a sample contains the assessment of all raters.
Next, point estimates \(\hat{K}_b\) are calculated for every sample.
The point estimates are sorted by size and are given by \(\hat{K}_B = ( \hat{K}_{[1]}, ...,\hat{K}_{[B]})\).
For the boundaries of the confidence interval (CI) at an error level of \(\alpha = 0.05\) the \(\frac{\alpha}{2}\) and \(1-\frac{\alpha}{2}\) percentiles of \(K_B\) are selected:

\begin{equation}
CI_{Bootstrap} = [\hat{K}_{B\frac{\alpha}{2}};\hat{K}_{B(1-\frac{\alpha}{2})}]
\end{equation}

Our two measures of IAA, the percentage agreement and Krippendorff's \(\alpha\),
indicate discrepancies between the different educational experts.
It follows that we have to be careful when generalizing our results to a larger readership.
A result that is repeatedly found in NLG studies~\cite{van2019best}. This is because language is multifaceted, and readers interpret the generated questions differently depending on prior knowledge and linguistic preferences~\cite{amidei2018rethinking}. IAA does not measure this multifacetedness. Moreover, in our case, it also does not capture the slightly different learning scenarios anticipated by the experts and the different ratings that accompany them.

Therefore, while IAA is important in NLG studies to assess the generalizability to a larger readership, one cannot draw direct conclusions about the quality of the generated questions from it.
It does not affect any conclusions about the generated questions' average quality. Thus, interesting observations can be derived from the reported results for our research question.

\begin{table}[tbp]
\caption{Generation statistics on different books. The mean (\(M_{q}\)), standard deviation (\(SD_{q}\)) and the median (\(Mdn_{q}\)) indicate the respective statistical value per section. As example, on average we generated 2.38 questions for each of the 60 sections in the biology textbook. }
\label{table_question_count_distribution}
\centering
\begin{tabularx}{0.5\textwidth}{l l l l l l l }
\hline
Book & \#sect & \#para & \#quest & \(M_{q}\) & \(SD_{q}\) & \(Mdn_{q}\)\\
\hline
overall & 325 & 1989 & 865 & 2.66 & 3.61 & 2 \\
anatomy & 96 & 451 & 335 & 3.49 & 4.14 & 2\\
biology & 60 & 285 & 143 & 2.38 & 3.16 & 2\\
chemistry & 44 & 312 & 118 & 2.68 & 4.75 & 1\\
physics & 43 & 348 & 62 & 1.44 & 2.44 & 0\\
psychology & 53 & 325 & 135 & 2.55 & 3.09 & 2\\
sociology & 29 & 268 & 72 & 2.48 & 2.43 & 2\\
\hline
\end{tabularx}
\end{table}

\begin{figure*}[tb]
\centering
\includegraphics[width=0.75\textwidth]{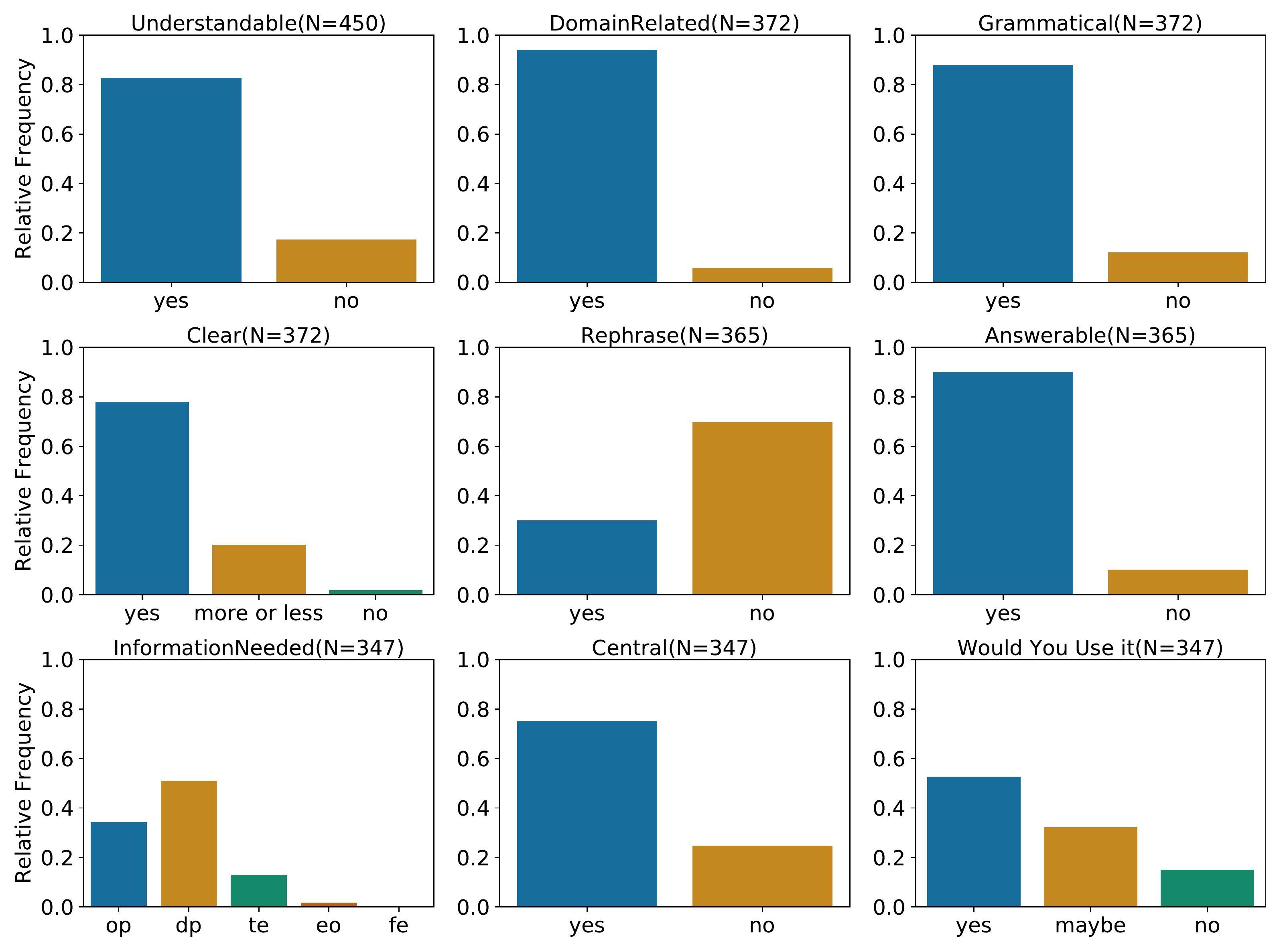}
\caption{Answer distributions in the evaluation study based on all three experts' annotations. The \(N\) indicates how often the given evaluation category was applicable. Hence, \(N\) decreases four times from category one of the annotation scheme (upper left corner) to category four of the annotation scheme (lower right corner). For the information need frequencies the following abbreviations are used: one position in the text (op), different positions in the text (dp), text + external knowledge (te), external knowledge only (eo) and feelings and beliefs of the readers (fe).}
\label{fig_evaluation_results}
\end{figure*}

\begin{table*}[tbp]
\caption{Examples of questions rated useful for teaching (example 1,2,3) or incomprehensible (example 4,5). In case the definition classification correctly identified a definition, the definiendum is marked in italic. The good questions where annotated as useful for teaching, whereas the bad questions were filtered out as incomprehensible. }
\label{table_examples}
\centering
\begin{tabularx}{\textwidth}{l l X >{\hsize=0.38\hsize}X}
\hline
\# & Rating & Context Sentence & Generated Question \\
\hline
1 & good & \emph{Metabolism} is the sum of all anabolic and catabolic reactions that take place in the body (Figure 1.6). & What is Metabolism the sum of?\\
2 & good & Sociologists often study culture using the \emph{sociological imagination}, which pioneer sociologist C. Wright Mills described as an awareness of the relationship between a person's behavior and experience and the wider culture that shaped the person's choices and perceptions. & How do sociologists study culture?\\
3 & good & Besides desegregation, sociology has played a crucial role in many important social reforms, such as equal opportunity for women in the workplace, improved treatment for individuals with mental disabilities or learning disabilities, increased accessibility and accommodation for people with physical handicaps, the right of native populations to preserve their land and culture, and prison system reforms. & What is one way sociology has played a role in social reforms? \\
4 & bad & The cell stores energy in the synthesis \emph{( anabolism )} of ATP, then moves the ATP molecules to the location where energy is needed to fuel cellular activities. & Where do ATP molecules move the ATP molecules to?\\
5 & bad & Less blood circulating means reduced blood pressure and reduced perfusion (penetration of blood) to the brain and other vital organs. & What is another way to describe less blood?\\
\hline
\end{tabularx}
\end{table*}

\subsection{Dataset Statistics}

An educational AQG that generates only one or two questions for a whole book is probably not considered useful by many.
Hence, we first discuss statistics on the number of questions generated and their distribution across the different books to estimate how many questions we can generate on average.
For the distributional statistics, we use the following terminology. One chapter comprises \(n\) sections (e.g. through subheadings) which again comprise \(m\) paragraphs divided by line separation.
Every generated question is related to a single paragraph and we report statistics on the section level (see Table \ref{table_question_count_distribution}).

We have 325 sections encompassing 1989 input paragraphs extracted from the first three chapters of the six input books.
Sections are of diverse length with a average word count of \(M_{para}= 505,28, Mdn_{para}=332\) and a standard deviation of \(SD_{para}=587.97\) words.
The longest sections are in the sociology textbook \(M_{soc} = 781.55\) and the shortest sections in the anatomy textbook \(M_{ana}=369.52\).
In general, this is indicative of different writing styles and textual structures in different domains.

On average, we were able to generate \(M_{quest}=2.67, Mdn_{quest}=2,00\) questions per section with a standard deviation of \(SD_{quest}=3.61\) questions.
The domain with the least questions generated per section is physics (\(M_{quest}=1,44, Mdn_{quest} = 0 \)).
The physics book frequently uses, formalized, non-natural language definitions such as mathematical equations.
Hence, it seems logical that the approach will generate fewer questions.
Interestingly, the generator performs better on the chemistry domain \(M_{quest}=2.68, Mdn_{quest} = 1\) regardless of the chemical formulas used. However, the book includes summary sections rephrasing the knowledge where the AQG performs well on.
Most of the generated questions start with \emph{``What is''} (\(53\%\)) followed by \emph{``What are''} (\(13\%\)) and \emph{``What does''} (\(12\%\)).
In general, around \(90\%\) of questions start with the word \emph{``What''}, followed by questions starting with \emph{``How''} in \(1\%\) of all cases.

\subsection{Expert Study Results}

The distributions of the annotations are shown in Fig. \ref{fig_evaluation_results}.
The results are averaged over all annotations.
As mentioned previously, the annotation scheme is hierarchical, and evaluation item marked as \emph{``not applicable''} are ignored for further analysis.
Consequently, the number of samples is lower for higher-order evaluation items. To allow a straightforward interpretation of the percentual values reported, we report the percentage relative to the remaining questions first, followed by the absolute percentage.
We first discuss the linguistic characteristics of the generated questions, followed by their answer characteristics and pedagogical usefulness.

To begin with, the generated questions are usually of high linguistic quality with \(83\%\) of questions being understandable and the majority of the understandable questions also being error-free (\(88\%/73\%\) total).
Most of these questions have been perceived as clear (\(78\%/64\%\) total) or at least more or less clear (\(20\%/17\%\) total).
We show five prototypical examples of questions in Table \ref{table_examples}.
The first three questions exemplify good generation runs, whereas example four and five are typical generation errors.
Although the questions are understandable and of high linguistic quality, they often follow relatively simple patterns such as \emph{``what is X?''} (Table \ref{table_examples}; example 1).
Templates may generate similar questions.
However, the proposed approach is free of handcrafted templates and works without domain-specific or text-specific adaptations.
It also generates fluent questions even if the input sentences are complex (Table \ref{table_examples}; example 2).
Additionally, the neural generator's expressivity helps in case of imperfections in the domain agnostic definition classification or answer selection. In some cases, e.g. in example 3, context selection resulted in an invalid definition being selected. In example 3 the selected answer was \emph{``increased accessibility and accommodation for people with physical handicaps''}.
The generator correctly inferred that it was only one possible characteristic mentioned in the sentence and therefore adapted the question accordingly.
Consequently, although the question is not strictly tackling a definition, experts still found it valuable.
This finding is in-line with the general high fluency and naturalness of texts generated by modern neural network-based language models in general, as found in other studies\cite{ippolito2019human}.
In the incomprehensible rated questions, the neural generator sometimes started to repeat words in odd ways (Table \ref{table_examples}; example 4), a typical decoding artefact in neural text generation \cite{holtzman2019curious}.
Furthermore, while the generator recovers from some context selection and answer selection errors, this is not always the case. We also observed the generation of grammatically well-formed but semantically uninterpretable questions. Those have been rated as incomprehensible by the experts (Table \ref{table_examples}; example 5).

Next, according to the experts, the majority of the questions are related to the text (\(94\%/78\%\) total), and the evaluation suggests that most of the questions are answerable by students working with the text (\(89\% /72\%\) total) .
Selecting subclauses as answers for the neural generation thus provides enough valuable context for the generator to produce semantically valid questions related to the text.
Matching this consideration is example 5, where the answer selection failed due to an incorrectly constructed dependency graph, resulting in an unnatural question.
According to the experts, for answering the question, most of the time, knowledge directly obtained in one place (\(51\%/39\%\) total) or in multiple places (\(34\%/26\% \) total) in the text is sufficient.
Based on the implemented approach, it is not surprising that the answer can often be found in one place in the text.
The structure of typical textbook texts may explain the frequency of the \emph{multiple places} response option.
If texts define a concept's characteristics in multiple sentences chained by coreferences, our approach currently only finds the first sentence, because we do not resolve coreference chains.
Hence, if we construct a question asking for a concept's characteristics, the follow-up sentences to the initial definition will not be considered during question generation, but may still be valuable for answering the generated question.

Finally, rating the educational usefulness is probably the most challenging and subjective task in the annotation scheme.
We nevertheless assume that the corresponding evaluation items provide valuable trends describing the pedagogical usefulness of the generated questions.
They estimate the centrality of the subject of the question regarding the comprehension of the text and the perceived usefulness for teaching with the given text.
Regarding centrality, most of the questions annotated to the end, ask for a central piece of information (\(75\%/ 57\%\) total).
Accordingly, we assume that in order to answer most questions, learners must retrieve or activate central propositions from their mental model.
In the cognitive models of reading comprehension, the reactivation of propositions usually promotes text comprehension.
Consequently, the given good ratings for the respondent's centrality indicate the pedagogical value of the generated questions.

Furthermore, the majority of the questions annotated to the end, are perceived useful for teaching by our experts (\(53\%/41\%\) total).
Additionally, many questions were rated \emph{maybe useful} by the experts (\(32\%/25\%\) total) .
Hence \(66\%\) of all generated questions have a certain educational value.
Note that this evaluation item is highly subjective.
However, since all experts bring along teaching experience, one would assume that they reject any didactically useless question, unanimously.
Instead, the experts state that they consider using the generated questions without agreeing on specific questions.
Reporting so many useful questions is an unlikely outcome for questions without any pedagogical quality.
On the other hand, if the experts attribute a certain pedagogical quality to the questions, but would only use questions suiting their preferences or learning context, results similar to the reported may be expected.
Therefore, we would view the evaluation item's results on whether or not experts would use the generated questions in their teaching sessions with a mild optimism.
However, we know that this is only a rough approximation, and future work needs to test that in concrete learning scenarios with actual students and teachers.

\section{Discussion And Limitations}

\subsection{Discussion}

The results yield nuanced insights on the linguistic and pedagogical quality criteria defined in chapter four.
First, the linguistic quality of the generated questions is high.
With a few exceptions, the questions are easy to understand and free of language errors.
Furthermore, they are sufficiently clear, and in most cases, a rephrasing of them is unnecessary.
They frequently resemble questions generated by templates, but more complex questions are also constructed depending on the input.
Nevertheless, the naturalness and variety of the generated questions might be improved.
With regard to the given research question, it is clear that the extent to which we can generate educationally meaningful questions is bound by the linguistic appropriateness of the generated questions.
If the generated questions are hard to understand, learners will experience high cognitive load trying to decipher their meaning.
Consequently, their mental representations of the text will degrade.
Additionally, learners can probably ignore a few textually broken questions but lose trust in the system's capabilities if the quality is too low.
Therefore, our results are encouraging because the approach produces very few linguistically inappropriate questions.
However, it can be expected that the low variety of questions harms learners' motivation in the long run.
Future work may investigate paraphrase generation or more advanced text decoding algorithms to diversify the questions generated.

Second, the results show a clear trend that questions generated with the proposed AQG have high text relatedness.
They are most often domain related and answerable.
When constructing the answer, learners must mainly use the information found directly in the text.
Consequently, the generated questions mainly address propositions in the text-base of learners.
Strong textual reference is useful because we aimed to support the construction of a coherent text-base.
The reader's construction processes frequently fail in this respect, and without a coherent text-base, it is impossible to build a coherent situational model.
However, we have to keep in mind that questions on the text-base should be complemented with questions addressing the propositions and inferences stored in the situational model.
After learners have developed a certain level of understanding in their mental representations, such questions are better able to target their interlinking and to create cognitive conflicts and thus stimulate deeper learning processes \cite{rouet2002mining}.
Besides that, our questions could also already produce the testing effect, thus ensuring better recall of the information asked from memory.
Therefore, we hypothesize that the generated questions can already contribute to text comprehension of the information asked, but should be supported by deeper questions.

Third, the results regarding the questions' pedagogical quality indicate a trend towards mostly central and often directly usable questions.
Asking for central concepts needed for text comprehension is crucial.
Questions guide the learner's attention and thus asking about less relevant information may lead to ignoring the crucial chunks of information.
Consequently, questions scaffolding text comprehension must inquire information about central propositions in the text-base to become useful.
Furthermore, although highly subjective, experts seem to see some pedagogical value in our questions and rate them often as useful.
In combination with the previously discussed theoretical considerations based on the various ratings and their impact on the learners' mental representations, there is evidence that our questions often have a certain degree of pedagogical value.

To conclude the discussion, the exploratory evaluation study conducted indicates that most of our questions are linguistically appropriate and possess a certain pedagogical foundation.
Therefore, we assume that we can generate questions improving text comprehension to a certain extent.
The exact extent of this depends on the learning goal, the text content and the readership. In future work, it may be evaluated in more extensive hypothesis-testing studies in concrete learning scenarios.
The systems investigated in such studies should consider that some useless questions are still generated in the worst case.
Therefore, it might be advisable to choose an author-focused study approach, so that authors can already sort out useless questions.

\subsection{Limitations of the Evaluation Study}

We would like to point out the following limitations of our research.
First, we opted for an explorative expert evaluation study.
Hence, the reported results are harder to interpret in the context of other studies investigating automatic question generation.
However, it has been argued that most automatic metrics such as BLEU \cite{papineni2002bleu} which have been used to compare such systems, are ill-suited for the task \cite{liu2016not,VanderLee2019} due to their low correlation with actual human judges.
Hence, a direct comparison of AQG systems without human evaluation has little value.
Furthermore, we phrase the AQG problem differently than related work.
Whereas previous studies usually define question-worthy information implicitly through their algorithms \cite{chen2019comparative,du2017identifying,rudian2020educational}, the proposed design is guided by the observation that definitions constitute pedagogically valuable information.
Consequently, a direct comparison on the question level is difficult, even when conducting human evaluation due to the approaches' different focus.
A fair comparison involves latent variables such as learning outcome or motivational factors whose investigation requires study designs with many learners.
Through the results reported, the value of conducting such an evaluation in the future becomes clear.

Second, the experts carrying out the evaluation study had no linguistic training and achieved modest IAA for most evaluation items.
Therefore, it may be argued that the reported data does not generalize well to a large readership.
We have chosen educational experts and not linguistics as annotators because the final goal is to use the generated questions in educational settings.
Thus, educational experts are closer to the final usage scenario.
We are aware of the potential problems with low IAA.
However, although the agreement between experts is not high, there are clear trends in most items' quality.
Furthermore, the use of IAA in text generation is limited, since good texts are highly variable, and the interpretation of the texts will always include subjective factors \cite{amidei2018rethinking}.
Consequently, future work should rather focus on learner-centred studies than on increasing experts IAA.

Third, our evaluation only encompasses three chapters from six different books, from six different domains.
Although we tried to include a diverse set of domains and books, it is difficult to say how well our results generalize to unseen texts.
The amount of data annotated is typical for NLG studies and puts us in the same range as other NLG studies \cite{VanderLee2019}.
The books are diverse in style, written by different authors and from different domains.
Hence, we are optimistic that our results generalize to other higher educational texts.

\section{Conclusion}

In this work, we presented a novel educational AQG. 
We evaluated its linguistic and pedagogical appropriateness in an explorative expert study using educational textbooks in various domains.
Unlike previous works, the AQGs content selection is guided by explicit pedagogical considerations and accepts complete textbooks as input.

The evaluation results indicate a high linguistic quality of the generated questions. They are fluent to read and are answerable.
Thus, the linguistic quality of the generated questions seems not to be the main challenge in educational question generation even when applied to arbitrary texts.
The pedagogical quality seems challenging to judge generally. The annotators frequently disagree on whether or not a question is usable, similar to other studies \cite{amidei2018rethinking,horbach2020linguistic}.
The reported evaluation results nevertheless hint a positive trend for the pedagogical appropriateness of the generated questions.
Furthermore, theoretical considerations concerning text comprehension support this conclusion.
Given the combined evaluation results, the questions seem to target the readers' text-bases' central propositions.
Thus we argue that the generated questions have at least some pedagogical value.
Based on these positive findings, further research in concrete learning scenarios is now needed to determine what constitutes a good question during higher educational reading comprehension.

\ifCLASSOPTIONcaptionsoff
  \newpage
\fi

\bibliographystyle{IEEEtran}
\bibliography{paper}

\begin{IEEEbiography}[{\includegraphics[width=1in,height=1.25in,clip]{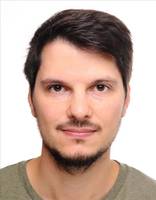}}]{Tim Steuer}
studied educational technologies at Saarland University.  Since the end of 2019, he works as a research assistant in the Knowledge Media research group at TU Darmstadt's  Multimedia Communications Lab (KOM). His research interests are centred on the intersection between educational technologies, human-computer-interaction and natural language processing.
\end{IEEEbiography}

\begin{IEEEbiography}[{\includegraphics[width=1in,height=1.25in,clip]{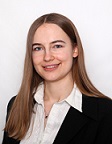}}]{Anna Filighera}
Anna Filighera graduated from the Technical University of Darmstadt with a Master's degree in computer science in 2019. Currently, she works as a research assistant at the Multimedia Communications Lab (KOM). Her research focuses on automatic knowledge assessment and feedback generation. 
\end{IEEEbiography}

\begin{IEEEbiography}[{\includegraphics[width=1in,height=1.25in,clip]{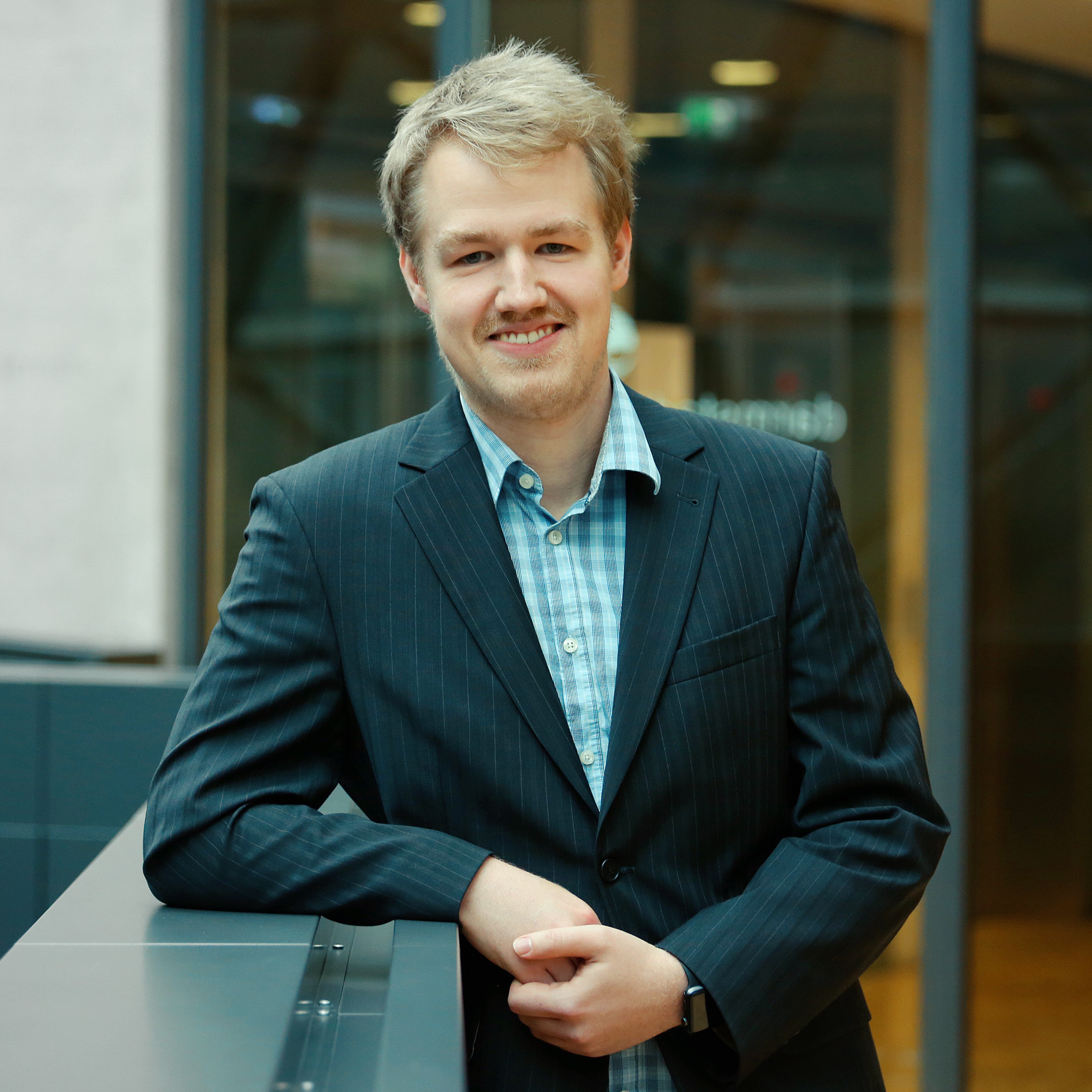}}]{Tobias Meuser}

studied business informatics at Fernuniversit{\"a}t Hagen and informatics at TU Darmstadt. Since the end of 2016, he is working as a research assistant in the research group "Distributed Sensing Systems" at KOM. Since begin of 2020, he is the head of the "Adaptive Communication Systems" Group at KOM. His research interest is centered around approximation techniques for communication networks, especially for network monitoring.

\end{IEEEbiography}

\begin{IEEEbiography}[{\includegraphics[width=1in,height=1.25in,clip]{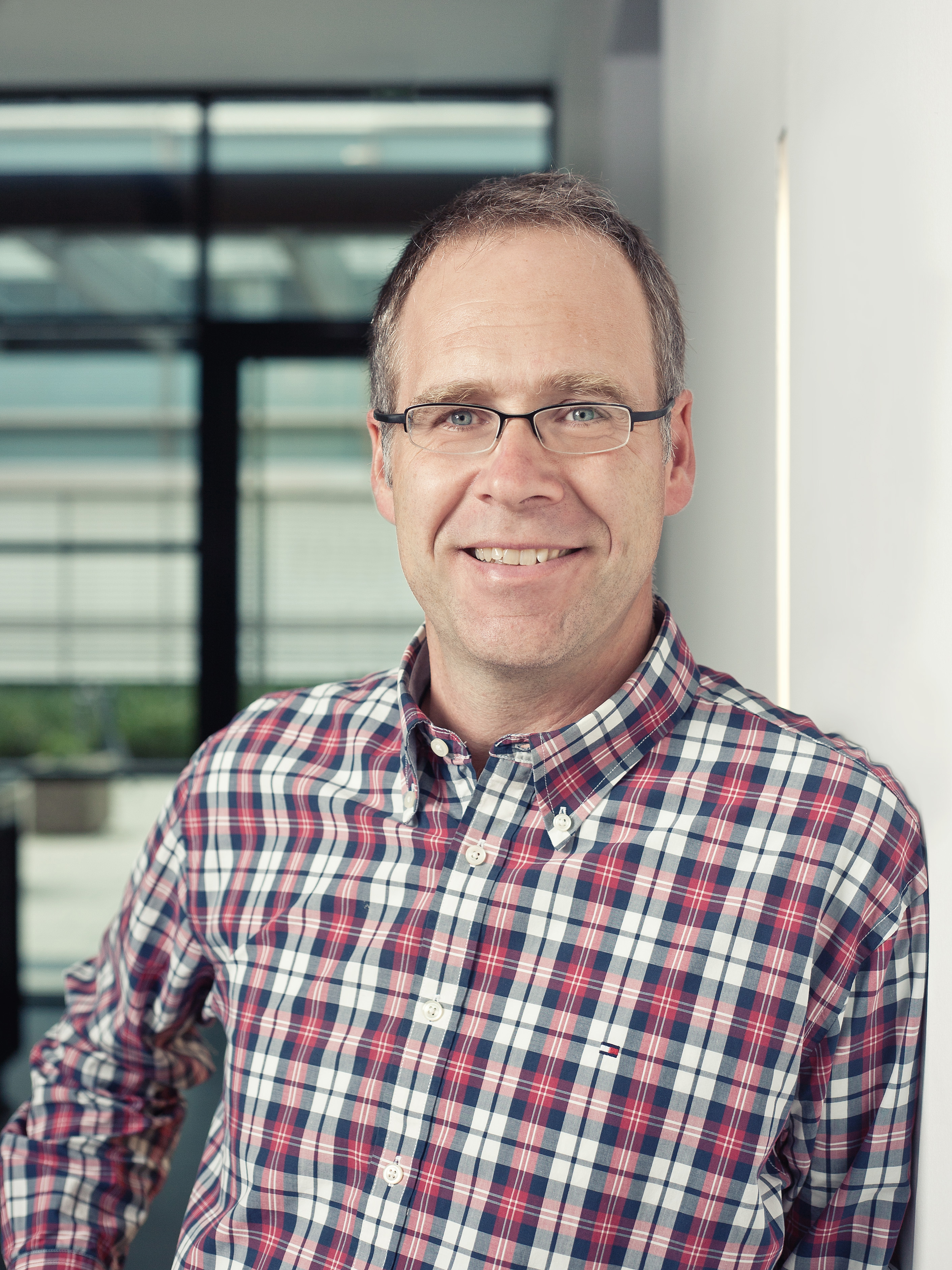}}]{Christoph Rensing} received the PhD degree in 2003, with a thesis on a policy-based access con-trol architecture for the multi-service Internet. At Technical University of Darmstadt he does research in technology enhanced Learning with a focus on adaptive learning technologies and learning analytics. He has published more than 180 papers in national and international conference proceedings and journals.
\end{IEEEbiography}
\end{document}